\documentclass[a4paper,10pt]{article}

\usepackage[utf8]{inputenc}
\usepackage[a4paper,left=2.5cm,right=2.5cm,top=2.5cm,bottom=2.5cm]{geometry}
\usepackage{natbib}
\usepackage[hidelinks]{hyperref}
\usepackage{graphicx}
\usepackage{bbm}
\usepackage{amsmath}
\usepackage{array}
\usepackage[font=small,labelfont=bf]{caption}
\usepackage{xspace}
\usepackage{pythonhighlight}

\usepackage{fancyhdr}

\newcommand{\pelenet}{\texttt{PeleNet}\xspace}
\newcommand{\nxsdk}{\texttt{NxSDK}\xspace}

\title{PeleNet: A Reservoir Computing Framework for Loihi}
\author{Carlo Michaelis}
\date{24. November 2020}

\begin{document}

\maketitle

\begin{center}
    Note: This is a draft
\end{center}

\begin{abstract}
\noindent High-level frameworks for spiking neural networks are a key factor for fast prototyping and efficient development of complex algorithms.
Such frameworks have emerged in the last years for traditional computers, but programming neuromorphic hardware is still a challenge.
Often low level programming with knowledge about the hardware of the neuromorphic chip is required.
The PeleNet framework aims to simplify reservoir computing for the neuromorphic hardware Loihi.
It is build on top of the NxSDK from Intel and is written in Python.
The framework manages weight matrices, parameters and probes.
In particular, it provides an automatic and efficient distribution of networks over several cores and chips.
With this, the user is not confronted with technical details and can concentrate on experiments.

\end{abstract}

\section{Introduction}

% Introduction and motivation
Several different neuromorphic hardware chips have been developed in recent years \citep[reviewed by][]{schuman2017survey, young2019review, rajendran2019low}.
All of them promise to be a key factor in future neuroscientific research as well as technological developments in artificial intelligence.
The main benefit of neuromorphic systems is their low power consumption and speed \citep{rajendran2019low}.
Shown for Loihi for example from \cite{tang2019spiking}.
This advantage of brain inspired hardware comes with a solution to the von Neumann bottleneck \citep{backus1978can}.
While novel neuromorphic hardware becomes more and more powerful, algorithms for such hardware systems are still in an early stage.
A specific field of spiking neural network algorithms, which can be used for neuromorphic hardware, is reservoir computing.
For details about reservoir computing I refer the reader to the literature \citep{jaeger2001echo, maass2002real, jaeger2007echo, schrauwen2007overview, lukovsevivcius2012reservoir, goodfellow2016deep}.

% Introduce functionality of Loihi
Here, I focus on the neuromorphic hardware chip Loihi \citep{davies2018loihi}.
The chip is digital and includes a current-based (CUBA) leaky integrate-and-fire (LIF) neuron.
A chip contains $128$ cores and each core time-multiplexes $1024$ compartments.
In addition, every chip contains three conventional x86 CPUs.
Several chips can be used in parallel on one board.
The parameters of the synapses and the compartments can be adapted, but the neuron model itself is fixed.
A single compartment neuron can be extended to a multi compartment neuron which comes with the cost of less available neurons.
Intel provides a software development kit, the so-called \nxsdk which is written in \texttt{Python} and already allows higher-level programming of the chip \citep{lin2018programming}.
In addition, \texttt{C} scripts can be used to run code on the x86 CPUs of the chip.
The \nxsdk allows to define \textit{compartment prototypes} and \textit{compartment groups} which can be combined to \textit{neuron prototypes} and \textit{neurons}.
Using \textit{connection prototypes} and \textit{connections} these compartments and neurons can be interconnected using a connection matrix.
Spikes can be injected using \textit{spike generators} and several different types of \textit{probes} can be defined.
Finally, \textit{learning rules} can be used to make connections in the reservoir plastic.

\begin{figure}[!t]
    \centering
    \includegraphics[width=\textwidth]{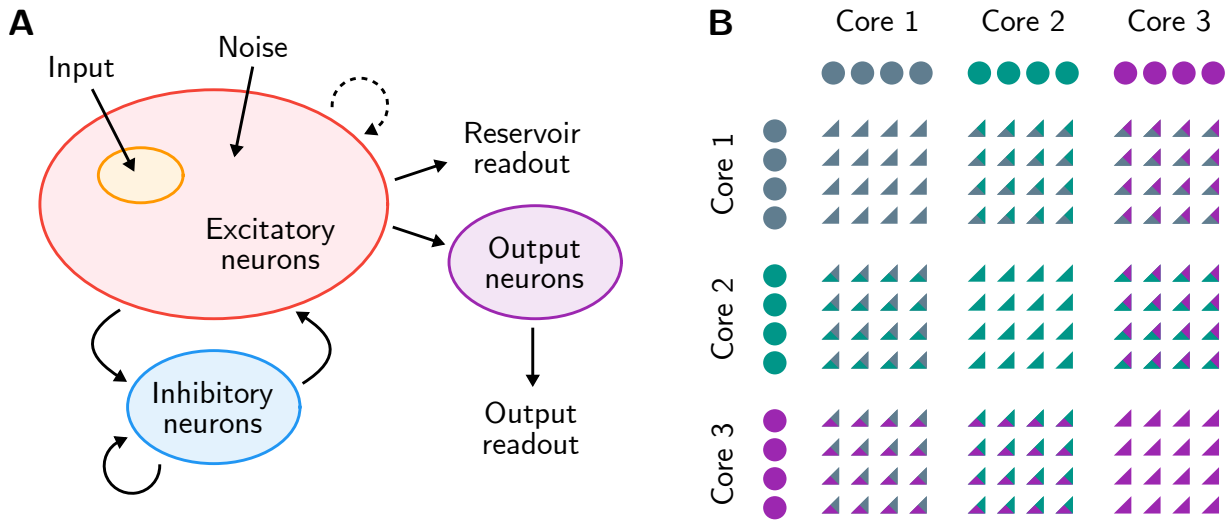}
    \caption{\textbf{(A)} The reservoir network consists of a pool of excitatory neurons (red) and a pool of inhibitory neurons (blue). Those pools are connected within and in-between. The excitatory neurons can be stimulated by an input (orange) and/or noise. Optionally a pool of output neurons (purple) can be defined. The spiking data can be read out from the reservoir itself and/or from the output neurons. \pelenet supports configuring a learning rule for the connections of the excitatory neurons (dotted arrow). \textbf{(B)} The example shows $12$ neurons which are distributed to three cores, where each core contains four neurons. Each core is color coded (grey, green, purple). Every triangle symbolises a potential connection between two neurons. The color of the triangle indicates which cores need to be interconnected to interconnect the related neurons. In this case, neurons spread over three cores require $9$ connection matrices.}
    \label{fig:1}
\end{figure}

% Limitation of nxsdk
However, the \nxsdk defines compartments for each core separately.
If bigger networks are used, it is necessary to split the connections manually to the Loihi cores.
Compartments between cores need then to be interconnected manually, which results in $n_{\text{conn}} = n_{\text{core}}^2$ connection matrices, if all potential connections should be possible.
For example, if we create $3$ compartment groups, distributed to $n_{\text{core}} = 3$ cores, and we want to interconnect all of them, we need to define $n_{\text{conn}} = 9$ connection weight matrices, which is illustrated in Figure~\ref{fig:1}B.
Note that this amount of matrices is necessary even for sparse reservoirs networks, since we do not want to exclude any possibility a priori.
In addition to handling these connection matrices, also probes can only be taken for each core individually.
Probing the whole network from the example above requires defining and handling $3$ probes for the compartment groups and $9$ probes for the connection weight matrices.
Note that this problem is only difficult in recurrent structures, especially in a reservoir where all neurons can potentially connect to each other.
In feed-forward structures, even in deep spiking neural networks \citep[already applied to several neuromorphic hardware systems, see e.g.][]{diehl2016conversion, schmitt2017neuromorphic, patino2020event, massa2020efficient}, this problem is less predominant since the layers are connected in series.
Multiple probes still need to be defined, but the connection matrices scale linear with the number of neuron groups and not quadratic as in reservoirs.

% Approach in PeleNet
The \pelenet framework was developed to solve the distribution of connections efficiently and to make it easy for the user.
In the framework, the experimenter only needs to define one connection matrix for every part of the network (e.g. for the reservoir or for the output layer).
After the simulation, the user gets usable probes for every part of the network.
In addition, \pelenet provides different distributions for initializing the connection weights, defining learning rules, creating standard plots, logging relevant computation steps and a collection of utils for calculating statistics and handling data.
Moreover, the framework is a whole new abstraction layer on top of the \nxsdk.
Compartments, connections and probes are defined implicitly and are controlled via parameters.
Due to its modular and object oriented architecture, the framework can easily be extended with additional functionality.
Here, I give a brief overview of the code structure and the main features.
The code is available on Github \footnote{PeleNet on Github: \url{https://github.com/sagacitysite/pelenet}} under the MIT license.

\section{Design and implementation}

\textit{Pele} is the goddess of volcanoes and fire in the Hawaiian religion \citep{nimmo1986pele, emerson2013pele}.
She has the control over lava and volcanoes and is inter alias in control of the volcano \emph{Loihi}.
The name of the \pelenet framework is an eponym of the goddess Pele.

The framework is build to allow experiments with reservoir networks on Loihi.
As shown in Figure~\ref{fig:1}A, \pelenet currently supports reservoir networks that follow Dale's law.
The reservoir contains a pool of excitatory and a pool of inhibitory neurons.
Those neuron pools are connected within and in-between.
Additionally an input, noise and an output is available.
The spiking data can be read out from the excitatory, inhibitory and output neurons.
Every experiment can contain one or multiple trials.
Figure~\ref{fig:3} shows an example with $10$ trials.
The spiking activity can optionally be reset after every trial.
With this, it is possible to simulate much faster, since is is not necessary to initialize the network again after every trial.

Programmatically, \pelenet consists of two main parts.
One part contains some helper functions and external libraries which are not available as a package.
This part is imported from the \pelenet framework internally, the user does not need to import modules from this part.
The other part consists of the \pelenet code itself which is imported and used by the user.

\subsection{Libraries}

The \texttt{lib} folder currently contains code to generate an anisotropic connectivity matrix and some helper functions and classes.
The code for initializing an anisotropic connectivity matrix is public on Github \footnote{Code for generating anisotropic connection weights on Github: \url{https://github.com/babsey/spatio-temporal-activity-sequence/tree/6d4ab597c98c01a2a9aa037834a0115faee62587}}.
The underlying principle was introduced in \citep{spreizer2019space} and used in \citep{michaelis2020robust} for generating robust robotic trajectories, using the \pelenet framework.
The \texttt{helper} folder contains custom exception functions for invalid parameters or invalid function arguments and a \texttt{Singleton} class to decorate classes in the \pelenet framework to make use of the singleton design pattern.

\subsection{PeleNet structure}

\begin{figure}[!t]
    \centering
    \includegraphics[width=0.7\textwidth]{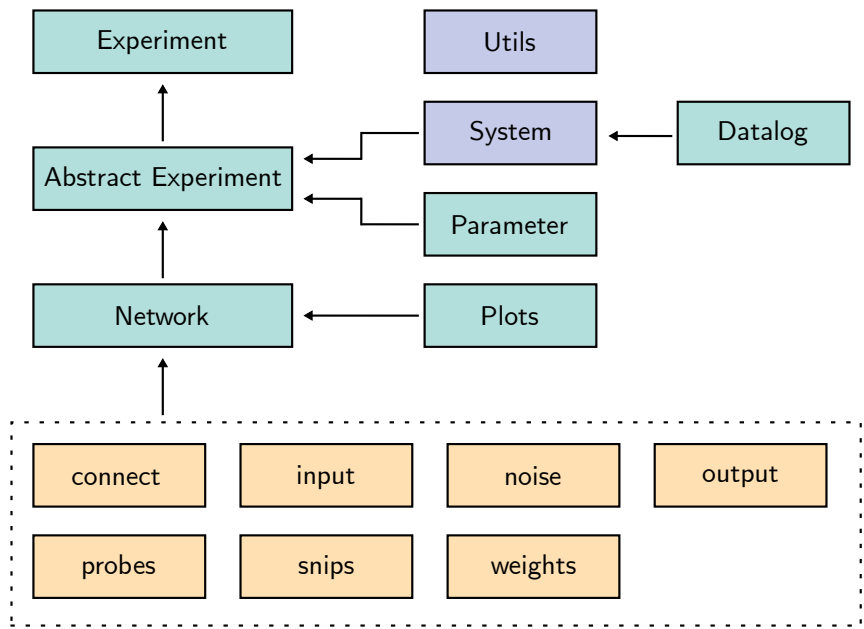}
    \caption{The code structure of the \pelenet framework. Classes have a green background, singletons are in purple and collections of methods for a class are yellow.}
    \label{fig:2}
\end{figure}

The central entity of \pelenet is the \textit{experiment}, which inherits from an \textit{abstract experiment}.
In an experiment one or several \textit{networks} can be defined and used.
In the experiment, a \textit{parameter} set is defined and overwrites values of the default parameter object.
The \textit{parameters} are passed to every \textit{network}, when it is initialized by the \textit{experiment}.
In addition, every \textit{network} contains a \textit{plot} object, which has access to the data sets and probes of the simulation.
Passing data to one of the \textit{plot} methods is therefore in most cases not necessary.
Hence, the arguments in den \textit{plot} method shape the plots appropriately in size, limits, labels, colors, etc.
Two singleton objects are globally available to all other objects.
The \textit{system} singleton contains a \textit{datalog} object which logs all important steps in a log file.
It also logs the parameter set which was used for a particular experiment and basic plots and optionally data sets from this experiment.
The \textit{utils} singleton provides a bunch of methods to handle and evaluate data, like dimensionality reduction, smoothing, calculating the spectral radius of the weight matrix and different kind of statistics.
Figure~\ref{fig:2} gives an overview of the dependencies between the classes in \pelenet.

\subsection{Network}

The major component of the \pelenet framework is the \textit{network}.
It contains a collection of methods, distributed over several files.
Since this is the core of the framework, I will give some more details about the behavior in the following listing.

\paragraph{weights}

Methods in the \texttt{weights} file initialize the weight matrices for all parts of the network.
The basic weight matrices are for connecting the excitatory and inhibitory parts of the reservoir.
Weights can be initialized using constant values or a log-normal or normal distribution.
In addition, a 2D topological anisotropic weight matrix can be initialized. 
All weight matrices are stored sparsely in a compressed sparse row (CSR) format.

\paragraph{connect}

Takes a weight matrix for every part of the network (e.g. one weight matrix for the whole reservoir), splits it in parts (called chunks in the framework), distributes these parts to the cores and interconnects them to each other.
It is currently still necessary to define the number of neurons that should be used per core as a parameter.
For an efficient distribution we need to consider two aspects.
First, every core time-multiplexes up to $1024$ neurons, the less neurons we simulate on every core, the faster the simulation will be.
Second, the more cores are used, the more connection matrices are required, which slows down the initialization of the network.
For an optimal run-time performance we should therefore reduce the number of neurons per core as much as possible and for an optimal initialization performance we should use as much neurons per core as possible.
It is currently up to the user to choose best fitting values for this trade-off, but in most cases it is probably the fast simulation time which is desired.
Later versions of the framework may allow the user to choose a preferred method and handles the distribution of neurons automatically.

\paragraph{input}

Adds different types of inputs to the network.
All of them base on the \textit{spike generators} from the \nxsdk.
Currently topological inputs (in case of a 2D-network), noisy inputs (leave-n-neurons out in every trial), sequences of inputs and varying input positions per trial are supported.
Topological inputs define a square of stimulated neurons in the excitatory layer of the reservoir at a defined position.
Noisy inputs stimulate a specific number of reservoir neurons, but in every trial some few neurons are left out such that the input differs slightly in every trial.
A sequence of inputs are multiple input regions which are stimulated in a row within one trial such that relations between them can be learned (e.g. with spike-timing-dependent plasticity).
An example for a input sequence is shown in Figure~\ref{fig:3}.
Finally, it is possible to define separate input regions, where one input region is randomly chosen in every trial.

\paragraph{noise}

Noise is currently generated by random inputs from \textit{spike generators} which are connected to randomly chosen neurons in the network.
Future implementations will probably make use of random changes in the current of a synapse or the membrane voltage of the neuron.

\paragraph{output}

Adds output neurons to the reservoir.
Currently the only available output is a pooling layer, which was used in \cite{michaelis2020robust}.
The pooling layer was used for a faster read out and performed regularization for the anisotropic network.

\paragraph{probes}

Contains several methods to define and process probes.
First, probes are defined for every core and every connection matrix chunk.
Second, after a successful simulation, probe data are post-processed and stacked together to complete and useful data sets.
Note that the output of a connection weight probes is in CSR format again such that this matrix can directly be used as an initial connection matrix for another simulation.

\paragraph{snips}

Handles small \texttt{C}-scripts that run on the x86 cores of the Loihi chips (so called SNIPs).
These scripts are located in the \texttt{pelenet/snips} folder.
Currently a reset SNIP is available that resets the membrane voltages after a trial.
It is important to note that the plasticity is currently not stopped while resetting the membrane voltages, which can cause problems when the $uk$ variables ares used.
But this feature is under development and will probably be added soon.

\begin{lstlisting}[
    style=mypython,
    frame=none,
    captionpos=b,
    label=code:experiment,
    float=t,
    numbers=left,
    caption={Defining an experiment in the \texttt{pelenet/experiments} folder. The \pyth{defineParameters} method is required. Lifecycle methods are optional (not shown). In addition, the experiment can be extended by custom methods for e.g. data evaluation or visualization.}
]
# Import abstract experiment
from ._abstract import Experiment

"""
@desc: An experiment with a sequential input, trained over several trials
"""
class SequenceExperiment(Experiment):

    """
    @desc: Define parameters for this experiment
    """
    def defineParameters(self):
        return {
            # Experiment
            'seed': 1,  # Set fixed random seed
            'trials': 10,  # Number of trials
            'stepsPerTrial': 60,  # Number of simulation steps for every trial
            # Neurons
            'refractoryDelay': 2, # Refactory period
            'voltageTau': 100,  # Voltage time constant
            'currentTau': 5,  # Current time constant
            'thresholdMant': 1200,  # Spiking threshold for membrane potential
            # Network
            'reservoirExSize': 400,  # Number of excitatory neurons
            'reservoirConnPerNeuron': 35,  # Number of connections per neuron
            'isLearningRule': True,  # Apply a learning rule
            'learningRule': '2^-2*x1*y0 - 2^-2*y1*x0 + 2^-4*x1*y1*y0 - 2^-3*y0*w*w',
            # Input
            'inputIsSequence': True,  # Activates sequence input
            'inputSequenceSize': 3,  # Number of input clusters in sequence
            'inputSteps': 20,  # Number of steps a trace input is active
            'inputGenSpikeProb': 0.8,  # Probability of spike for the input generator
            'inputNumTargetNeurons': 40,  # Number of neurons activated by the input
            # Probes
            'isExSpikeProbe': True,  # Probe excitatory spikes
            'isInSpikeProbe': True,  # Probe inhibitory spikes
            'isWeightProbe': True  # Probe weight matrix at the end of every trial
        }
\end{lstlisting}

\subsection{Experiments}

An \textit{experiment} inherits from an \textit{abstract experiment} and is created in the \texttt{pelenet/experiments} folder.
The \textit{abstract experiment} inherits again from the \texttt{ABC} package, which allows defining abstract methods.
The \pyth{defineParameters} method in the \textit{abstract experiment} class is implemented as an abstract method that is then necessary in every \textit{experiment}, otherwise an exception is thrown.
The \textit{abstract experiment} also provides some default functionality, which can optionally be overwritten.
It initializes all necessary objects for the experiment and contains a default build process.
Also the execution of the simulation follows a default behavior.
In both cases, it is preferred to control the behavior of the \textit{experiment} via parameters instead of overwriting methods.
If the parameters do not cover the wanted behavior, it is suggested to make use of the available lifecycle methods.
Available lifecycle methods are:

\begin{itemize}
    \item \texttt{onInit}: Called after the experiment was initialized.
    \item \texttt{afterBuild}: Called after all network parts are connected (i.e. weight matrix, inputs, outputs, noise, probes).
    \item \texttt{afterRun}: Called after the simulation has finished and all data are post-processed.
\end{itemize}

Only if the parameters and the lifecycle methods are not sufficient to solve the intended behavior it is suggest to overwrite the \texttt{build} and \texttt{run} methods of the \textit{abstract experiment}.

\begin{figure}[!t]
    \centering
    \includegraphics[width=\textwidth]{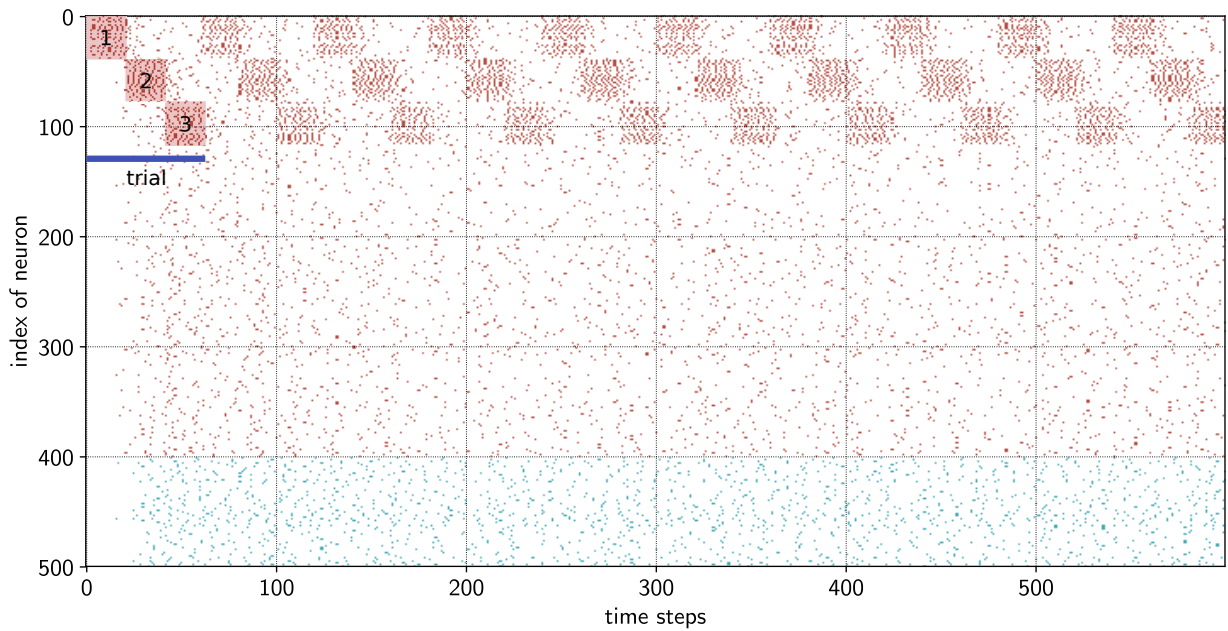}
    \caption{Example for a spike raster plot that shows inputs and trials. The experiment was performed with $10$ trials and a input sequence with three inputs.}
    \label{fig:3}
\end{figure}

In practice, \texttt{Jupyter} notebooks are used to quickly evaluate the results of an \textit{experiment}.
It is suggested to use \texttt{Jupyter} notebooks only for visualization of the results and for prototyping.
Code for the experiment should be included in the defined \textit{experiment}.
An example of a simple \textit{experiment} is shown in Code Listing \ref{code:experiment}.
In Code Listing \ref{code:jupyter} the experiment is used.
The plotted spike train from the Code Listing \ref{code:jupyter} (line $25$) is shown in Figure~\ref{fig:3}.

\subsection{Parameters}

\begin{lstlisting}[
    style=mypython,
    frame=none,
    captionpos=b,
    label=code:jupyter,
    float=t,
    numbers=left,
    caption={An example for running the experiment. For this Jupyter notebooks can be used. Parameters of the experiment can be overwritten to allow fast experimentation. At the end of the script an included plotting method is called to show spike trains. They are directly plotted in Jupyter and stored in the log folder related to this simulation.}
]
# Load pelenet modules
from pelenet.experiments.sequence import SequenceExperiment

# Overwrite default parameters from pelenet/experiments/sequence.py
parameters = {
    # Experiment
    'seed': 2,  # Change random seed
}

# Initialize experiment
# The name is optional, it is extended to the folder in the log directory
# The parameters defined above are handed over to the experiment object
exp = SequenceExperiment(
    name='random-network-sequence-learning',
    parameters=parameters
)

# Build the network (weight matrix, inputs, probes, etc)
exp.build()

# Run the network simulation, afterwards the probes are post-processed to nice arrays
exp.run()

# Plot spike trains of the excitatory and inhibitory neurons
exp.net.plot.reservoirSpikeTrain(figsize=(12,6))
\end{lstlisting}

The parameter system is a powerful tool of \pelenet for defining an experiment.
All \nxsdk functionalities, which are included in \pelenet, are covered by the parameter set.
It is not required to know anything about the \nxsdk at all, if the functionality provided by \pelenet is sufficient for the user.
The default parameters are split in three parts, parameters for the experiment, the system and derived parameters.
The system parameters cover information about e.g. the used Loihi board, settings for the \texttt{matplotlib} library, logging and paths.
The experiment parameters contain e.g. neurons, connections, inputs, outputs, noise, probes and also a learning rule for the reservoir.
Finally, the derived parameters are calculated from the system and experiment parameters and are useful for the framework or for the user.
Some parameters are sanity checked to avoid serious issues with false parameters.
These checks are constantly extended.
Note that the parameters are well documented in the \texttt{pelenet/parameters} file to understand their meaning, but it is not suggested to overwrite the parameters there.
For overwriting parameters the \pyth{defineParameters} method is available in the \textit{experiment}.
All parameter values which are defined in this method overwrite the default parameter set.
Derived parameters are calculated after the parameters are defined in an \textit{experiment}.

\section{Discussion}

% introduction
Currently the neuromorphic hardware community grows fast and it is probably only a matter of time until new or updated hardware systems will emerge.
The field of applications is very broad for such hardware, including the estimation of linear models, like LASSO \citep{shapero2014optimal, davies2018loihi}, non-parametric classification with k-nearest neighbor \citep{frady2020neuromorphic}, deep spiking neural network \citep{massa2020efficient} or reservoir networks \citep{michaelis2020robust}.
High-level libraries and frameworks are needed to cover these different specialized application areas.
While for deep spiking neural networks the \texttt{SNN Toolbox} \citep{rueckauer2017conversion, rueckauer2018conversion} is available for ANN to SNN conversions (also for Loihi) or \texttt{SLAYER} is available for training SNNs via backpropagation, reservoir network frameworks are still rare for neuromorphic computing.

% summary & strenghts
The here presented \pelenet framework simplifies the implementation of reservoir networks on the neuromorphic hardware Loihi.
The framework is an abstraction layer on top of the \nxsdk from Intel.
\pelenet allows an efficient distribution of the network over an arbitrary number of Loihi cores and chips.
Probes are combined to data sets, which can directly be used for further evaluations.
Parameters define the experiments.
The ``parameter approach'' allows an easy initialization process and keeps the experiments clear.
Finally, the plot systems already includes a bunch of default plots for reservoir computing.

% limitations and next steps
Beside its already existing features, the framework is still in an early stage.
Some functionalities of the Loihi chip are not provided yet.
This includes inter alia current and voltage noise, usage of the tag for learning rules and synaptic delays.
The learning rule covers the excitatory neurons in the reservoir, future version will also be able to apply a learning rule to output neurons.
In addition, the framework is not yet available as a \texttt{pip} python package and still requires a manual installation of dependencies.
Therefore one of the next steps will be to provide a package release of \pelenet.
It is also intended to add an optimization functionality which runs several experiments in order to find optimal parameters, according to a criterion.
The parameter system is already designed to support future optimization scripts.
Finally, it is planned to add unit and integration tests to make sure that simulations are performed correctly.
Currently four \texttt{Jupyter} notebooks are available as functional tests.

% conclusion
The aim of \pelenet is to speed up the implementation of reservoir computing experiments on Loihi.
In a more broad perspective it has even the potential to push the field of reservoir computing implementations on neuromorphic hardware in general.
Despite its early development stage, I am confident that the framework is already useful for computational studies.

\section{License and code availability}

The framework \pelenet is published under the MIT License and is therefore freely available without warranty. The code is available on Github\footnote{PeleNet on Github: \url{https://github.com/sagacitysite/pelenet}}. Contributions are highly appreciated.

\section{Acknowledgement}

I sincerely thank Dr. Christian Tetzlaff for providing me with funding for my doctoral thesis and his supervision. In addition, I want to thank Intel for providing the Kapoho Bay to our lab and ongoing support. Finally I want to thank Andrew B. Lehr, since he is always willing to discuss problems constructively and influenced the implementation of the anisotropic network significantly.

\bibliography{bib}{}
\bibliographystyle{apa}

\end{document}